\documentclass[runningheads,a4paper]{llncs}

\usepackage{amssymb,amsmath}
\usepackage{blindtext}
\usepackage{graphicx}
\usepackage[misc]{ifsym}
\usepackage{url}
\urldef{\mailsa}\path|{yxz772@case.edu}|   
\newcommand{\keywords}[1]{\par\addvspace\baselineskip
	\noindent\keywordname\enspace\ignorespaces#1}
\graphicspath{ {Img/} }
\usepackage{ctable}

\newcommand{\E}{\mathbb{E}}
\usepackage{multirow}

\newcommand\blfootnote[1]{%
	\begingroup
	\renewcommand\thefootnote{}\footnote{#1}%
	\addtocounter{footnote}{-1}%
	\endgroup
}

\begin{document}
	
	\mainmatter  
	
	\title{Task Driven Generative Modeling for Unsupervised Domain Adaptation: \\Application to X-ray Image Segmentation}
	
\author{Yue Zhang$^{1,2}$\textsuperscript{(\Letter)} \and Shun Miao$^{1}$\and Tommaso Mansi$^{1}$  \and Rui Liao$^1$}
%

	\authorrunning{Y. Zhang et al.}
	\titlerunning{Task Driven Generative Modeling}
\institute{$^{1}$Medical Imaging Technologies, Siemens Healthineers Technology Center, \\
	Princeton, NJ 08540, USA \\
	$^{2}$Department of Mathematics, Applied Mathematics and Statistics, \\
	Case Western Reserve University, Cleveland, OH 44106, USA \\
	\mailsa
}
	%
	%
	
	\maketitle

	\begin{abstract}
			\vskip -0.05in
		Automatic parsing of anatomical objects in X-ray images is critical to many clinical applications in particular towards image-guided invention and workflow automation. Existing deep network models require a large amount of labeled data. However, obtaining accurate pixel-wise labeling in X-ray images relies heavily  on skilled clinicians due to the large overlaps of anatomy and the complex texture patterns. On the other hand, organs in 3D CT scans preserve clearer structures as well as sharper boundaries and thus can be easily delineated. In this paper, we propose a novel model framework for learning automatic X-ray image parsing from labeled CT scans. Specifically, a Dense Image-to-Image network (DI2I) for multi-organ segmentation  is first trained on X-ray like Digitally Reconstructed Radiographs (DRRs) rendered from 3D CT volumes. Then we introduce a Task Driven Generative Adversarial Network (TD-GAN) architecture to achieve simultaneous style transfer and parsing for unseen real X-ray images. TD-GAN consists of a modified cycle-GAN substructure for pixel-to-pixel translation between DRRs and X-ray images and an added module leveraging the pre-trained DI2I  to enforce segmentation consistency. The TD-GAN framework is  general and can be easily adapted to other learning tasks. In the numerical experiments, we validate the proposed model on 815 DRRs and 153 topograms.  While the vanilla DI2I without any adaptation fails completely on segmenting the topograms, the proposed model does not require any topogram labels and  is able to provide a promising average dice of $85\%$ which achieves the same level accuracy of supervised training (88\%).\blfootnote{\noindent\textbf{Disclaimer: }This feature is based on research, and is not commercially available. Due to regulatory reasons its future availability cannot be guaranteed.}
	\end{abstract}

	\section{Introduction}
	
	Semantic understanding of anatomical objects in X-ray images is critical to many clinical applications, such as pathological diagnosis, treatment evaluation and surgical planning. It serves as a fundamental step for computer-aided diagnosis and can enable  intelligent workflows including organ-based autocollimation, infinite-capture range registration, motion compensation and automatic reporting. In this paper, we study one of the most important problems in semantic understanding of X-ray image, i.e., multi-organ segmentation.
	
	While X-ray understanding is of great clinical importance, it remains a very challenging task, mainly due to the projective nature of X-ray imaging, which causes large overlapping of anatomies, fuzzy object boundaries and complex texture patterns. Conventional methods rely on prior knowledge of the procedure (e.g., anatomical motion pattern from a sequence of images \cite{zhu2009dynamic}) to delineate anatomical objects from X-ray images. Modern approaches utilize deep convolutional networks and have shown superior performance \cite{ronneberger2015u}. However, they typically require a large amount of pixel-level annotated training data. Due to the heterogeneous nature of X-ray images, accurate annotating becomes  extremely difficult and time-consuming  even for skilled clinicians. On the other hand, large pixel-level labeled CT data are more accessible. Thousands of X-ray like images, the so-called Digitally Reconstructed Radiographs (DRRs), are generated from labeled CTs and used in \cite{albarqouni2017x} to train an X-ray depth decomposition model. While using automatically generated DRRs/labels for training has merits, the trained model  cannot be directly applied on X-ray images due to their appearance 
	Generalization of image segmentation models trained on DRRs to X-ray images requires \textit{unsupervised domain adaptation}. 
	While many effective models \cite{bousmalis2016domain,tzeng2017adversarial}  have been studied, most of them focus on feature adaptation which naturally suits for recognition and detection. However, segmentation task desires pixel-wise classification which requires delicate model design and is substantially different. Recently, pixel-level adaptation models \cite{bousmalis2017unsupervised,zhu2017unpaired} have been proposed which utilize generative adversarial networks and achieve promising results on image synthesis and recognition. Still, continuing study on image segmentation especially for medical applications remains blank.

\begin{figure}[t]
	\centerline{
		\includegraphics[width=1\textwidth]{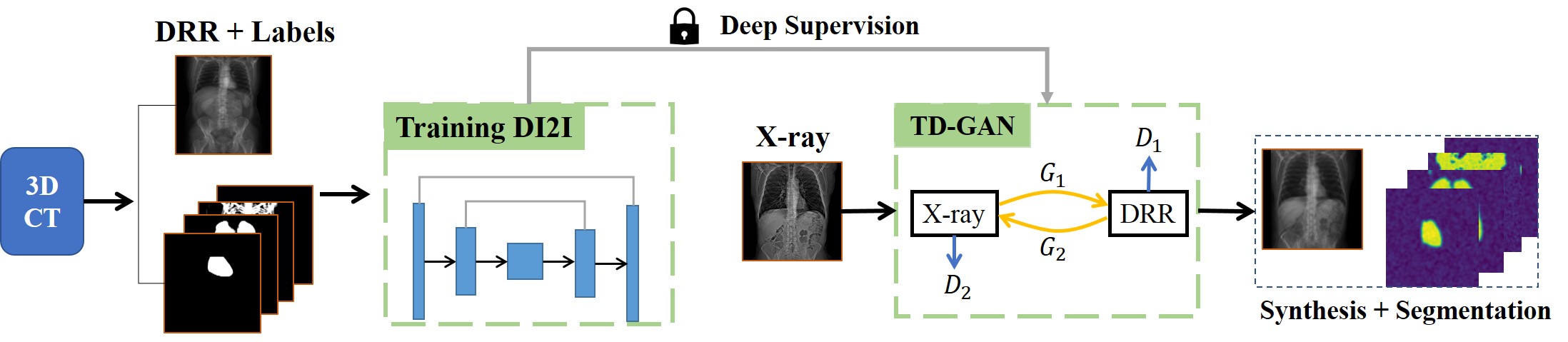}
	}
	\caption{Overview of the proposed task driven generative model framework.} \label{fig:intro}
\end{figure}

	In this paper, we present a novel model framework to address this challenge. Specifically, we first create DRRs and their pixel-level labeling from the segmented pre-operative CT scans. A Dense Image-to-Image network (DI2I) \cite{huang2017densely,jegou2017one} is then trained for multi-organ (lung, heart, liver, bone) segmentation over these synthetic data. Next, inspired by the recent success of image style transfer by cycle generative adversarial network (cycle-GAN) \cite{zhu2017unpaired}, we introduce a task driven generative adversarial network (TD-GAN) to achieve simultaneous image synthesis and automatic segmentation on X-ray images, see Figure \ref{fig:intro} for an overview. We emphasize that the training X-ray images are \textbf{unpaired} with previous DRRs and are totally \textbf{unlabeled}. TD-GAN consists of a modified cycle-GAN substructure for pixel-to-pixel translation between  DRRs and X-ray images. Meanwhile,  TD-GAN incorporates the pre-trained DI2I to obtain deep supervision and enforce consistent performance on segmentation. The intuition behind TD-GAN is indeed simple: we transfer X-ray images in the same appearance as DRRs and hence leverage the pre-trained DI2I model to segment them. Furthermore, the entire transfer is guided  by the segmentation supervision
	 network. 

	The contributions of our work are: 1) We propose a novel model pipeline for  X-ray image segmentation from unpaired  synthetic data (DRRs). 2) We introduce an effective  deep architecture TD-GAN for simultaneously image synthesis and segmentation without any labeling effort necessary from X-ray images. To our best knowledge, this is the first end-to-end framework for unsupervised medical image segmentation. 3) The entire model framework can be easily adjusted for unsupervised domain adaptation problem where labels from one domain is completely missing.  4) We conduct numerical experiments and demonstrated the effectiveness of the proposed model with over 800 DRRs and 150 X-ray images.

	\section{Methodology}
	\subsection{Problem Overview}
	In this paper, our goal is to learn a multi-organ segmentation model on  unlabeled X-ray images using  pixel-wise annotated DRRs data. Furthermore, these images are not paired, which means they are not taken from the same group of patients. We collected X-ray images that contain targeted organs such lung, heart, liver and bone (or a subset of them). The DRRs of the same region of interest (ROI) are generated by placing 3D labeled CT volumes in a virtual imaging system that simulates the actual X-ray geometry.   Meanwhile, the pixel-level labeling of DRRs are generated by projecting 3D CT labels along the same trajectories. While most public datasets for multi-organ segmentation only consists of tens of cases, our dataset covers a richer variety of scanning ranges, contrast phases as well as morphological differences. In the next subsections, we first train a DI2I on the DRRs and then adapt it to the X-ray images with TD-GAN to provide deep segmentation supervision during the image synthesis.
	

	\subsection{Dense Image to Image Network for Segmentation on DRRs}
\begin{figure}[t]
	\centerline{
		\includegraphics[width=1\textwidth]{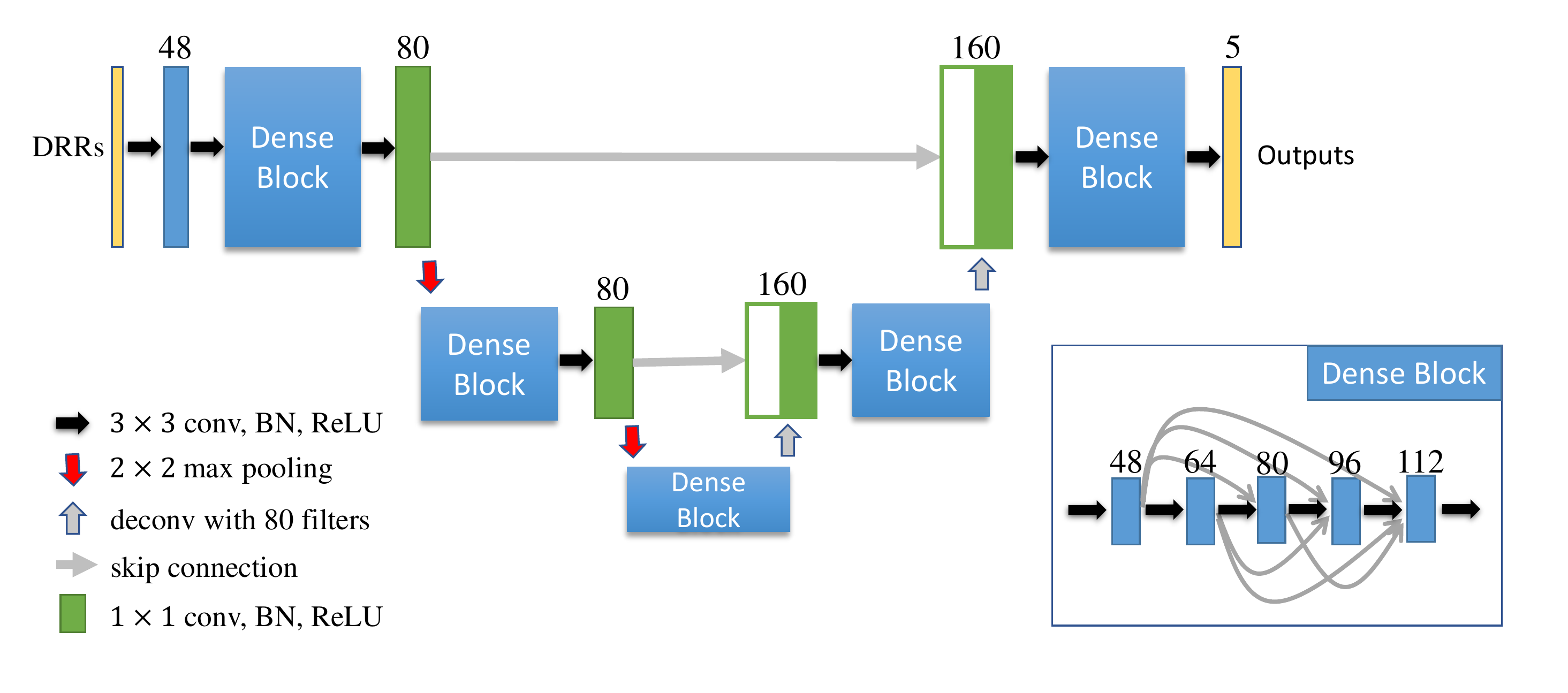}
	}
	\vskip -0.15in
	\caption{Illustration of the DI2I for segmentation on DRRs.  } \label{fig:denseunet}
	\end{figure}

	We  train a Dense Image-to-Image network (DI2I) on the labeled DRRs data. As is depicted in Figure \ref{fig:denseunet}, the network employs an encoder-decoder UNet structure with dense blocks \cite{ronneberger2015u,huang2017densely,jegou2017one}. The network consists of dense blocks which are generalizations from ResNets \cite{he2016deep} by iteratively concatenating all feature  outputs in a feed-forward fashion.  This helps alleviating the vanishing gradient problem and thus can obtain a deeper model with higher level feature extraction.  Despite these appealing properties, empirically we found that it achieved superior performance than classical UNet. 
	 The final output feature map has five channels which consists of a background channel $x_0$ and four channels $x_1,...,x_4$ corresponding to the four organs. By doing so, we alleviate the challenge of segmenting overlapped organs and simplify the problem into binary classifications. We further use a customized loss term which is a weighted combination of binary cross entropies  between each organ channel and background channel, 
	\begin{equation}
		\mathcal{L}_{seg} = -\sum_{i=1}^{4} w_i(y_i\log(p_i) + (1-y_i)\log(1-p_i))
	\end{equation}\label{eq:seg_loss}
	where $y_i$ is the ground truth binary label map for each organ and $p_i$ is calculated as $ \exp(x_i)/(\exp(x_0)  + \exp(x_i))$ for $i=1,2,3,4$.
	
	\begin{figure}[t]
	\centerline{
		\includegraphics[width=1\textwidth]{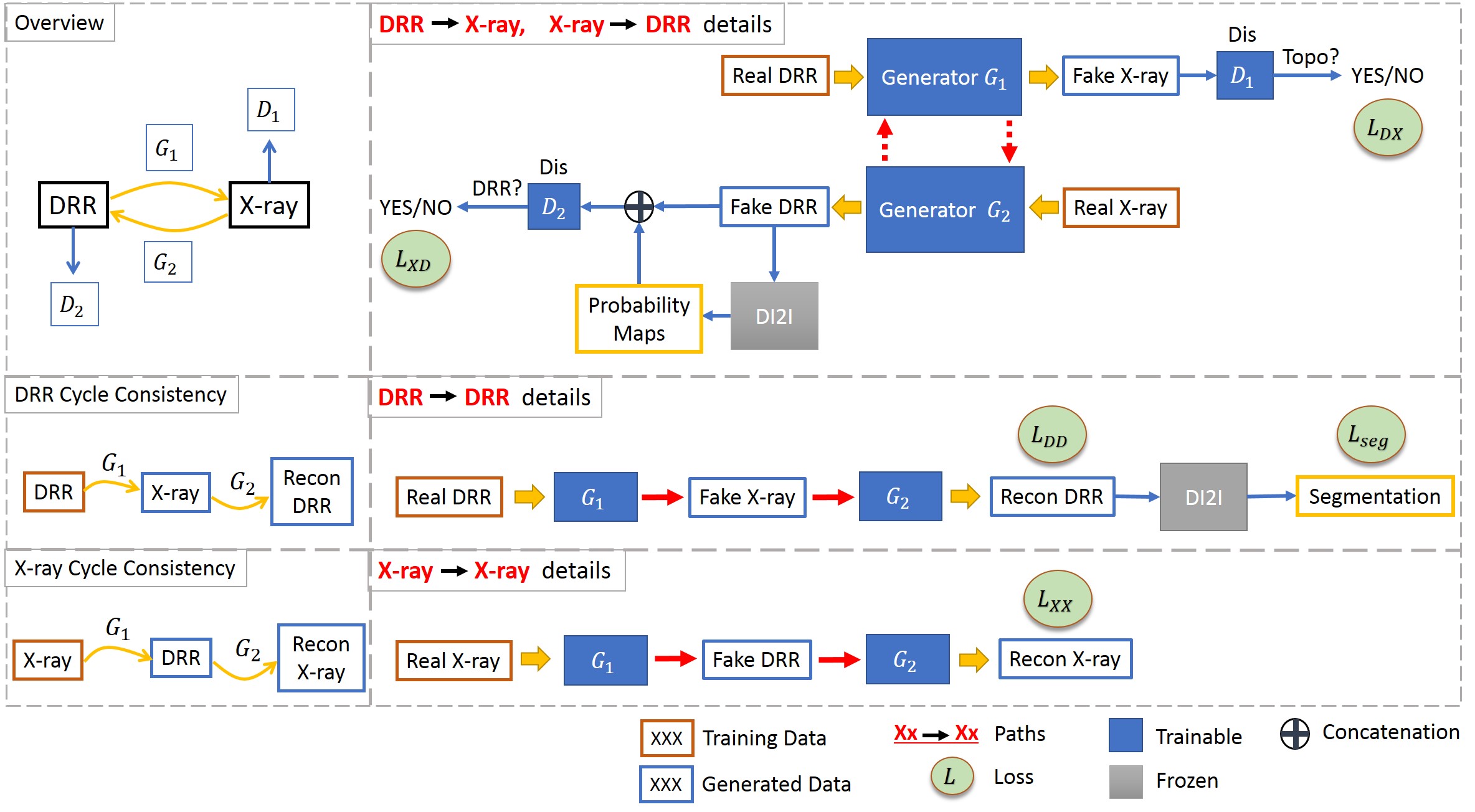}
	}
	\caption{Proposed TD-GAN architecture. Real X-ray images and DRRs are passed through 4 different paths for simultaneously synthesis and segmentation. } \label{fig:architecture}
		\vskip -0.1in
	\end{figure}	
	
	\subsection{Task Driven Generative Adversarial Networks (TD-GAN)}
	We now describe the main deep architecture of TD-GAN. The network has a modified cycle-GAN \cite{zhu2017unpaired} sub-structure with add-on segmentation supervisions. As is depicted in  Figure \ref{fig:architecture}, the base structure consists of two generators $G_1, G_2$ and two discriminators $D_1, D_2$ to achieve pixel-to-pixel image synthesis. The generators try to synthesize images in the appearance of the other protocol while the discriminators need to distinguish the generated images (\textbf{fake}) against the reals. When the cyclic training converges to optimal,  the network will be able to transfer X-ray images to DRRs through $G_2$ and transfer back through $G_1$. This also holds for DRRs. However, image synthesis only serves for a general purpose of appearance transfer and is not segmentation focused. Important prior knowledge such as organ boundaries, shapes and local variations are not carefully treated and could possibly lose during the synthesis process.  We therefore add supervision modules to enforce the segmentation consistency, that is, during the transfer process we  require the X-ray images that \textit{not only}  have the appearance of DRRs  \textit{but also} can be segmented by the pre-trained DI2I network. This is done by introducing conditional adversarial training on the translation of X-ray images and cycle segmentation consistency on DRRs. We detail this later on. 

	The proposed TD-GAN involves 4 different paths that transfer and reconstruct images between the two protocols: real DRR $\rightarrow$ fake X-ray, real X-ray $\rightarrow$ fake DRR,  real X-ray $\rightarrow$ reconstructed X-ray and real DRR $\rightarrow$ reconstructed DRR. Different losses are proposed for each path. We discuss the meaning of these paths in a top-down topological order that is shown in Figure \ref{fig:architecture}.	We denote the data distribution $d\sim p_d$ and $x \sim p_x$ for DRRs and X-ray images. Our main contribution comes in the segmentation driven losses in path real X-ray $\rightarrow$ fake DRR and real DRR $\rightarrow$ reconstructed DRR.
	
	\textbf{Real DRR $\rightarrow$ Fake X-ray}.	Given a real DRR image, the generator $G_1$ tries to produce the corresponding image in the appearance of X-ray images. The discriminator $D_1$ will need to distinguish the generated fake X-ray image and the real. Since we do not have any paired X-ray images with the DRRs, this real X-ray image is randomly selected from the training dataset. A successful generation from $G_1$ will confuse $D_1$ to make the wrong prediction. The loss function involved is a standard GAN loss for image style transfer,
	\[
	\mathcal{L}_{DX} := \E_{t\sim p_x}\left\lbrace  \log\left[ D_1(x) \right] \right\rbrace  + \E_{d\sim p_d}\left\lbrace  \log\left[ 1 - D_1(G_1(d))\right] \right\rbrace . 
	\]
	 
	\textbf{Real X-ray $\rightarrow$ Fake DRR}. The other generator $G_2$ will produce a fake DRR to challenge $D_2$. We could also randomly select a real DRR in the training of $D_2$. However, this is suboptimal since the labels of DRRs will be unused which contain important organ information such as size, shape and location and are crucial to our final segmentation task.  Inspired by the conditional GANs \cite{mirza2014conditional}, we leverage the pre-trained DI2I to predict the organ labels on the fake DRRs. The fake DRRs combined with their predicted labels are then fed into $D_2$ to compare with the real pairs. Therefore $D_2$ needs to not only distinguish the fake DRRs and the reals but also determine whether the image-label pairs are realistic. To confuse $D_2$, the generator $G_2$ will particularly focus on the organs of interest during the image transfer. We hence will obtain a more powerful generator in the task of segmentation. Finally, to make the involved loss function differentiable, we only obtain the predicted probability map from the DI2I and do not binarize them. Denote $U(\cdot)$ as the pre-trained DI2I, we have the following loss function for this path,
	\begin{equation}\label{eq:T1}
		\begin{aligned}
			\mathcal{L}_{XD} := &\ \E_{d\sim p_d}\left\lbrace  \log\left[D_2(d|U(d))\right] \right\rbrace \\
			& + \E_{x\sim p_x} \left\lbrace  \log\left[1 - D_2(G_2(x)|U(G_2(x)))\right]\right\rbrace. 
		\end{aligned}
	\end{equation}
	
	We remark that the pre-trained DI2I is frozen during the training of TD-GAN, otherwise the supervision will be disturbed by the fake DRRs. Furthermore, TD-GAN can be easily adapted to other tasks by replacing $U(\cdot)$.  
	
	\textbf{Real X-ray $\rightarrow$ Reconstructed X-ray}. The idea behind this path is that once we transferred a X-ray image to a fake DRR through $G_2$, we should be able to transfer this DRR back to X-ray image through $G_1$.  The final (reconstructed) X-ray image should be the same as the original one. This is called cycle-consistency. The corresponding loss function is calculated by $l_1$ distance,
	\[
	\mathcal{L}_{XX} := \E_{x\sim p_x} \left\lbrace \|G_1(G_2(x)) - x\|_1\right\rbrace
	\]
	
	\textbf{Real DRR $\rightarrow$ Reconstructed DRR}. Same argument also applies to the DRR reconstruction. Moreover,  we enforce the reconstructed DRR to maintain cycle segmentation consistency by adding a segmentation loss as a regularization term. The implies that we would like that the reconstructed DRR is not only close to the original but maintain the same segmentation performance under DI2I. The two losses involved in this path are as follows,
	\begin{align}
		\mathcal{L}_{DD} & := \E_{d\sim p_d} \left\lbrace \|G_2(G_1(d)) - d\|_1\right\rbrace, \nonumber \\
		\mathcal{L}_{seg} & := \mathcal{L}_{seg}, \label{eq:T2}
	\end{align}
	where the segmentation loss $\mathcal{L}_{seg}$ is the same as in equation (1).
	
	The total loss for TD-GAN is then a weighted summation of all the losses in the above paths. 
	 We demonstrate the effectiveness of these two losses in the experiments. We employ the same parameter setting as is suggested in \cite{zhu2017unpaired}. The generator $G_1$ and $G_2$ utilize a same Resnet20 structure. The discriminator $D_1$ and $D_2$ contains four consecutive convolutional layers with increasing number of filters from 64 to 512 and a final  output layer with sigmoid activation. 

	\section{Experiments and Results}
	
	In this section we validate our methodology on a dataset of 815 labeled DRRs and 153 topograms. The topograms are acquired  before the CT scan for isocentering and therefore co-registered with the CT. The CT scans are labeled pixel-wisely and the labels of topograms are generated in a same way as  the aforementioned DRRs masks. 	The co-registered CT is not used in training and the labels on topograms are used only for evaluation purpose. Our model is among the first approaches to address unsupervised domain adaptation for segmentation with unpaired data. We compare with the state-of-the-art image synthesis model cycle-GAN \cite{zhu2017unpaired} since both can be used in our problem except ours are task driven. To further show the effectiveness of the  deep architecture and the proposed losses, we also compare our models with TD-GAN adversarial (TD-GAN-A)  and TD-GAN reconstruction segmentation (TD-GAN-S) by enabling either one of the conditional adversarial loss (\ref{eq:T1}) and the cycle segmentation consistency loss (\ref{eq:T2}).
	
	We first train a DI2I for multi-organ segmentation on the labeled DRRs. A standard 5 fold cross-validation scheme is used to find the best learned weights. We evaluate the dice score on the testing dataset, summarized as follows (mean $\pm$ std): lung 0.9417 $\pm$ 0.017, heart 0.923 $\pm$ 0.056, liver 0.894 $\pm$ 0.061 and bone 0.910 $\pm$ 0.020. All the experiments are run on an 12GB NVIDIA TITAN X GPU.
\vskip -0.1in
	\begin{figure}
	\centerline{
	\includegraphics[width=0.22\textwidth]{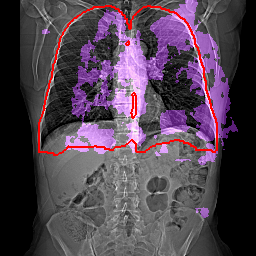}
	\includegraphics[width=0.22\textwidth]{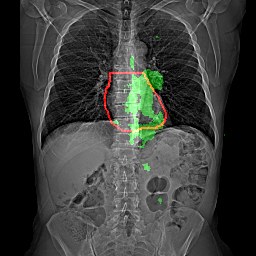}
	\includegraphics[width=0.22\textwidth]{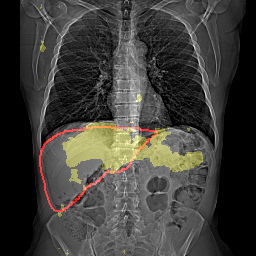}
	\includegraphics[width=0.22\textwidth]{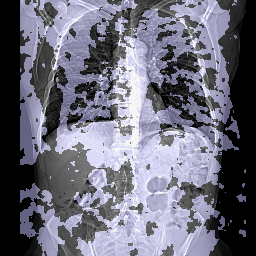}
} \vskip 0.05in
	\centerline{
		\includegraphics[width=0.22\textwidth]{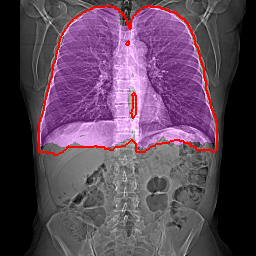}
		\includegraphics[width=0.22\textwidth]{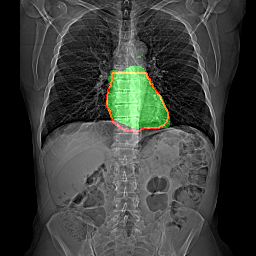}
		\includegraphics[width=0.22\textwidth]{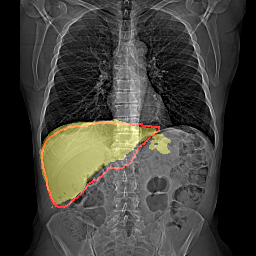}
		\includegraphics[width=0.22\textwidth]{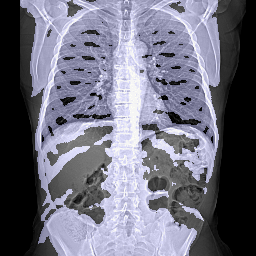}
	} 
	\caption{Visualization of segmentation results on topograms (bottom)  against direct application of DI2I (top). The red curves stands for the boundary of the ground truth. The colored fill-in parts are the predictions by TD-GAN and DI2I.}\label{fig:seg_topo}
	\vskip -0.1in
\end{figure}	
	
	Next we load the pretrained DI2I into TD-GAN with weights all frozen and train the model to segment the topograms. We use all the DRRs as well as 73 topograms for training, 20 topograms for validation and 60 for testing. 
	 We visualize one example and compare the proposed method against vanilla setting (where the DI2I is tested directly on topograms) in Figure \ref{fig:seg_topo}.  For better visualization purpose, we only show the ground the truth labeling for lung, heart and liver. The numerical results are summarized in table \ref{tb:topo}. To better understand the performance of our model, we also train the DI2I on the topograms using their labels under the same data splitting setting, listed as Supervised in table \ref{tb:topo}. 
	
	While the direct application of the learned DI2I on topograms fails completely, it can be seen that our model significantly improved the segmentation accuracy and even provided the same level of accuracy compared with the supervised training with labeled topograms. Compared with the cycle-GAN which only performs image style transfer, both the partially task driven nets TD-GAN-A  and TD-GAN-S can improve the performance. Furthermore, the final TD-GAN combines the advantages of all the three models and achieves the best.\vskip -0.05in
	\begin{table}
		\centering
		\caption{Average Dice results   of segmentation on topograms. } \label{tb:topo}
		\begin{tabular}{c|c|c|c|c|c|c}
			\hline\noalign{\smallskip}
			Objects & Vanilla & CGAN     & TD-GAN-A   & TD-GAN-S   & TD-GAN    & Supervised    \\ 
			\noalign{\smallskip}
			\hline
			\noalign{\smallskip}
			Bone   & 0.401   & 0.808  & 0.800 & 0.831 & \textcolor{blue}{\textbf{0.835}}  & 0.871 \\ 
			Heart  & 0.233     & 0.816& 0.846 & 0.860 & \textcolor{blue}{\textbf{0.870}} & 0.880 \\  
			Liver  & 0.285     & 0.781& 0.797 & 0.804 &\textcolor{blue}{\textbf{0.817}}  & 0.841 \\  
			Lung   & 0.312     & 0.825 & 0.853 & 0.879 & \textcolor{blue}{\textbf{0.894}} & 0.939 \\  \hline
			mean   & 0.308    & 0.808 & 0.824 & 0.844 & \textcolor{blue}{\textbf{0.854}}  & 0.883\\ \hline
		\end{tabular}\vskip -0.1in
	\end{table}
	
	\section{Discussions and Conclusions}
	
	In this paper, we studied the problem on multi-organ segmentation over totally unlabeled X-ray images with labeled DRRs.	Our model leverages a cycle-GAN substructure to achieve image style transfer and carefully designed add-on modules to simultaneously segment organs of interest.  The proposed model framework is general. By replacing the DI2I with other types of supervision networks, it can be easily adapted to many scenarios in computer-aided diagnosis such as prostate lesion classification, anatomical landmark localization and abnormal motion detection. We leave this for the future direction.
	\vskip 0.1in

	\bibliographystyle{splncs}
	\bibliography{ref2}

\end{document}